\documentclass{article}

\usepackage[final]{nips_2017}

\usepackage{amsmath}
\usepackage[utf8]{inputenc} 
\usepackage[T1]{fontenc} 
\usepackage{hyperref} %
\usepackage{url} 
\usepackage{booktabs} %
\usepackage{amsfonts} 
\usepackage{nicefrac}
\usepackage{microtype} %
\usepackage{threeparttable}
\usepackage{enumitem}
\usepackage{etoolbox}
\usepackage{url}

\newcounter{inlineenum}
\renewcommand{\theinlineenum}{\alph{inlineenum}}
\newenvironment{inlineenum}
  {\unskip\ignorespaces\setcounter{inlineenum}{0}%
   \renewcommand{\item}{\refstepcounter{inlineenum}{\textit{\theinlineenum})~}}}
  {\ignorespacesafterend}

\title{\vspace{-0.3cm}On Evaluating and Comparing Open Domain Dialog Systems\vspace{-0.3cm}}

\author {
  Anu Venkatesh \thanks{Equal Contribution}, $^1$ Chandra Khatri \footnotemark[1], $^1$ Ashwin Ram,$^1$ Fenfei Guo\thanks{Work done during an internship at Amazon Alexa}, $^2$ \\
  \textbf{ Raefer Gabriel,$^1$ Ashish Nagar,$^1$ Rohit Prasad,$^1$ Ming Cheng,$^1$}\\
   \textbf{ Behnam Hedayatnia,$^1$ Angeliki Metallinou,$^1$ Rahul Goel,$^1$ Shaohua Yang,\thanks{Work done during an internship at Amazon Alexa} $^3$Anirudh Raju$^1$}\\
  $^1$Amazon Alexa, $^2$ University of Maryland, $^3$Michigan State University \\
  \texttt{\{anuvenk,ckhatri,ashwram,raeferg,nashish\}@amazon.com} \\
   \texttt{\{roprasad,chengmc,behnam,ametalli,goerahul,ranirudh\}@amazon.com} \\
  \texttt{fenfeigo@cs.umd.edu}, \texttt{yangshao@msu.edu} \\ 
\vspace{-1cm}
}

\begin{document}

\maketitle

\begin{abstract} \vspace{-0.0135\textheight}
    
    

      
    
    

Conversational agents are exploding in popularity. However, much work
remains in the area of non goal-oriented conversations, despite
significant growth in research interest over recent years. To advance
the state of the art in conversational AI, Amazon launched the Alexa
Prize, a 2.5-million dollar university competition where sixteen
selected university teams built conversational agents to deliver the
best social conversational experience. Alexa Prize provided the
academic community with the unique opportunity to perform research
with a live system used by millions of users. The subjectivity
associated with evaluating conversations is key element underlying
the challenge of building non-goal oriented dialogue systems. In this
paper, we propose a comprehensive evaluation strategy with multiple
metrics designed to reduce subjectivity by selecting metrics which
correlate well with human judgement. The proposed metrics provide
granular analysis of the conversational agents, which is not captured
in human ratings. We show that these metrics can be used as a
reasonable proxy for human judgment. We provide a mechanism to unify
the metrics for selecting the top performing agents, which has also
been applied throughout the Alexa Prize competition. To our knowledge,
to date it is the largest setting for evaluating agents with
millions of conversations and hundreds of thousands of ratings from
users. We believe that this work is a step towards an automatic
evaluation process for conversational AIs.
  

\end{abstract}\vspace{-0.02\textheight}

\vspace{-0.3cm}
\section{Introduction} \vspace{-0.01\textheight}
\vspace{-0.005\textheight}

Conversational interfaces have recently became a focal point in both academia and industry research for several reasons, such as: \begin{inlineenum}
\item Rise of digital assistants like Amazon Alexa, Cortana and Siri,
\item Presence of universal chat platforms with socialbots like Facebook
  Messenger and Google Allo,
\item Advances in Machine learning and natural
  language understanding (NLU) systems, and
\item Introduction of NLU services such as Amazon Lex. 
\end{inlineenum} “Chatbots” are one specific type of conversational interface with no
explicit goal other than engaging the other party in an interesting or
enjoyable conversation. 
While modern chatbots have progressed since ELIZA~\citep{weizenbaum1966eliza}, current state-of-the-art systems are still a
long way from being able to have coherent, natural conversations with
humans~\citep{levesque2017common}. Alexa Prize was established to
advance the state of the art in this area and bring current research
to a production environment with hundreds of thousands of users. One of
the main challenges faced by researchers is the lack of a
good mechanism to measure the performance due to lack of explicit objective for open domain conversations. The Turing
Test~\citep{turing1950computing} is a well-known test that can
potentially be used for chatbot evaluation. However, we do not believe
that Turing Test is a suitable mechanism to evaluate chatbots for the following reasons:
\begin{itemize}[leftmargin=*,topsep=0pt,noitemsep]
\item \textbf{Incomparable elements}: Given the amount of knowledge an
  AI has its disposal, it is not reasonable to suggest that a human
  and AI should generate similar responses. A conversational agent may
  interact differently from a human, but may still be a good
  conversationalist.
  \item \textbf{Incentive to produce plausible but low-information
    content responses}: If the primary metric is just generation of
    plausible human readable responses, it is easy to opt out of the
    more challenging areas of response generation and dialogue
    management. It is important to be able to source interesting and
    relevant content while generating plausible responses.
\item \textbf{Misaligned objectives:} The goal of the judge should be
  to evaluate the conversational experience, not to attempt to get the
  AI to reveal itself.
\end{itemize}
A well-designed evaluation metric for conversational agents that
addresses the above concerns would be useful to researchers in this
field. Due to the expensive nature of human based evaluation
procedures, researchers have been using automatic machine translation (MT)
metrics such as BLEU~\citep{papineni2002bleu} or text summarization
metrics such as ROUGE~\citep{papineni2002bleu} to evaluate
systems. But as shown by~\cite{liu2016not}, these metrics do not correlate well with human
expectations. ~\cite{serban2015survey} recently did a survey on
available datasets for building and evaluating conversational dialogue
systems, which illustrated another problem - there is a lack of high
quality, open-ended, freely available conversational datasets. This is
concerning because these datasets are used to compare different
proposed metrics by researchers in the field. Those which exist
(e.g. Reddit, Twitter) have issues with quality, number of turns,
tracking context and multiple agent conversations.  The Alexa Prize is
uniquely poised in this regard as the conversation and evaluation
happen in real time through voice based interactions, with a rating provided immediately at the end
of the conversation by the user who actually had the interaction with the agent. The fact that the conversation is verbal is
important because people behave differently when talking vs
writing~\citep{redeker1984differences}. In the context of Alexa Prize,
the specific goal of simulating social conversation and the lack of a
commonly-agreed upon standard meaning for “chatbot” led to the use of
the term “socialbot” to describe the competition’s conversational
agents, which is a chatbot capable of interacting on a range of open domain conversational topics common in social conversation.

To evaluate the Alexa Prize socialbots, we developed a framework based
on engagement, domain coverage, coherence, topical diversity and
conversational depth. We show that these metrics correlate well with
human judgement by validating against more than hundreds of thousands of conversations. We believe this is the largest evaluation of conversational agents to date.

\vspace{-0.4cm}
\section{Related Work} \vspace{-0.01\textheight}
\vspace{-0.005\textheight}


Evaluation of dialogue systems is a challenging research problem, which
has been heavily studied but lacks a widely-agreed-upon
metric. However, there is significant previous work on evaluating goal-oriented dialogue systems. Some of the notable earlier works include
TRAINS system ~\citep{ferguson1996trains},
PARADISE~\citep{walker1997paradise}, SASSI~\citep{hone2000towards} and
MIMIC~\citep{chu2000mimic}. All of these systems involve some
subjective measures which require a human in the loop.  However, it is
easier to evaluate a task-oriented dialogue system because we can measure systems by successful completion of tasks, which is not the case with open-ended systems.

Automated metrics such as BLEU and METEOR~\citep{banerjee2005meteor},  are used for machine translation or ROUGE, which is used for text
summarization have been popular metrics to evaluate dialogue systems
as they can be easily calculated for a given dataset without the need
for human intervention. However, these metrics are primarily focused
on token-level overlap between surface forms. A valid and interesting
response to a statement in a conversation might have low semantic or
token level overlap with a reference response.~\cite{liu2016not} show
that these metrics show either weak or no correlation with human
judgements. 
There is work in MT and Natural Language Generation (NLG) fields which
studies correlation of BLEU with human judgement and highlights some
of its
shortcomings~\citep{graham2015accurate,espinosa2010further,cahill2009correlating}.~\cite{shawar2007different}
suggests a framework similar to ours based on dialogue efficiency,
dialogue quality and user satisfaction; however their work involves
small corpora and it is unclear how their framework will scale
to large datasets. There has also been work on learning scoring
models for evaluation of MT models in WMT evaluation
task~\citep{callison2011findings,bojar2016results}. Such models have
been used in both the MT~\citep{gupta2015reval,albrecht2007regression}
and chatbot domains~\citep{lowe2017towards,higashinaka2014evaluating}
for evaluation. While these models try to capture some
aspect of coherence, fluency or appropriateness of output, they are all
dependent on context and only perform well in a particular setting.
Such models can be components of a framework which compares
chatbots, training them can also be a challenge due to expensive data
requirements~\citep{lowe2017towards}.

More recently, adversarial evaluation, first proposed
by~\citep{bowman2015generating, adversarialTesting} is gaining traction.
~\cite{DBLP:journals/corr/LiMSRJ17} use adversarial learning in a
reinforcement learning (RL) setting for training and evaluation of
their dialogue generation model. In other directions,~\cite{graham2017can} try to use crowdsourced workers to evaluate
MT systems but there is no consensus on ``good human evaluation'' and
reliably getting such evaluations can be prohibitively expensive.
While some of these techniques apply to dialogue interfaces, general
responses based on conversational context are much more diverse, which
makes dialogue evaluation a harder problem. In this direction ~\cite {paradise1997}, proposed a PARADISE framework using which they decoupled the task requirements from the dialogue behavior and supported comparisons among dialogue strategies. 

\vspace{-0.4cm}
\section{Alexa Prize} \vspace{-0.01\textheight}
\vspace{-0.005\textheight}

The Alexa Prize (\cite{alexaPrize} ) is an annual university competition that was set up
with the goal of furthering conversational AI. Conversational AI is
hard for a multitude of reasons including but not limited to the need
for good free-form ASR, language understanding, dialog and context
management, language generation and personalization. University teams
participating in the Alexa Prize were tasked to build agents which can hold social conversations on popular topics and news
events in domains such as Politics, Sports, Entertainment, Fashion and
Technology. A user request to initiate a chat with Alexa is routed
into the Alexa Prize experience. One of the participating socialbots
is randomly launched anonymously and connected to the user to continue the
conversation. At the end of the conversation, the user is
asked to rate the conversation 
by answering the following question: "On a scale from 1 to 5 stars how do you feel about speaking with this socialbot again?" and leave a free-form feedback to the university team for improving socialbot quality. 
This setup allowed us to generate a large-scale human-evaluated conversational dataset with ratings.

There are two critical parts to the process of building an effective
model for any purpose: relevant data and an effective evaluation
strategy. Over the duration of the competition, users initiated over a
million conversations. The Alexa Prize enabled university teams to
access real user conversational data at scale, along with the
user-provided ratings and feedback. This allowed them to effectively make
improvements throughout the duration of the competition while being evaluated in real-time. Before the finals, we observed a 14.5\% improvement
in ratings across all of the socialbots (average rating increased from
2.76 to 3.16) and 20.9\% improvement across the 3 finalists (average
rating increased from 2.77 to 3.35). One unique aspect of this
conversational setup is that the user providing the rating for the
conversation is the person who engaged in the conversation itself. In
most prior work for non-goal oriented conversations, evaluations has been performed offline
by separate raters. The Alexa Prize offered a unique opportunity to
generate an evaluated dataset that represented the interactor’s point
of view on the experience. This dataset is critical to being able to
evaluate the effectiveness of the objective matrix that we propose in
this paper as an alternate automated mechanism to evaluate
conversations.

\vspace{-0.4cm}
\section{Evaluation Metrics}\label{eval} \vspace{-0.01\textheight}
\vspace{-0.005\textheight}

Some of the proposed metrics are based on topic identification; hence Section 4.1 provides insight on the topic extraction module. We list a set of metrics (Section 4.2 - 4.7) that can be computed to objectively evaluate and compare conversational agents.  These metrics were also validated against a dataset of hundreds of thousands of ratings and were found to be in-line with human evaluation. 


\vspace{-0.3cm}
\subsection{Extracting Topical Metrics} \label {topical-extraction} \vspace{-0.01\textheight}
\vspace{-0.003\textheight}
We trained a topic classification model using Deep Average Network (DAN) (Iyyer et al., 2015) on an Alexa internal dataset to identify topics within a conversation to enable computation of the above topic-based metrics. The classifier identified topical domains for any given user utterance or socialbot response into one of 26 predefined topical domains (Sports, Politics, etc.) with 82.4\% accuracy (obtained through 10 fold cross-validation). This model is used to obtain various topical metrics proposed in the following sections. In this paper, references to topics relate to topical keywords such and "obama", "nba" etc. and domains to the 26 topical domains referenced above.

\vspace{-0.3cm}
\subsection{Conversational User Experience (CUX)} \vspace{-0.01\textheight}
\vspace{-0.003\textheight}

Conversations with a socialbot can be significantly different from those with humans. The reasons noted below are potential contributors to the difference:

\vspace{-0.01\textheight}

\begin{itemize}[leftmargin=*,topsep=0pt,noitemsep]
\itemsep0em 
\item \textbf {Expectation:} The purpose for which someone may engage with a socialbot may be significantly different – e.g. some users expect accurate answers to questions, while others merely want a friendly presence to listen empathetically. The baseline expectations of a conversational agent’s capabilities also seemed to vary significantly among users.
\item \textbf {Behavior and Sentiment:} The potential lack of fear of affecting a relationship may lead to different degree of freedom of expression with agents.
\item \textbf {Trust:} There may be difference in opinion on how secure conversations with socialbots are, especially compared to humans. As a user builds trust in a system, they may begin to engage in ways indicating higher trust in the socialbot through opinion requests, conversation requests indicating a need for companionship etc.
\item \textbf {Visual Cues and Physicality:} Absence of visual cues and physical signals such as prosody, body langugage may impact the conversation content and direction.
\end {itemize}

It is hard to capture the above-mentioned metrics
numerically. To capture the subjectivity involved in evaluating the experience, we used Alexa user ratings. The high variability in ratings can be noted by the fact that for a range of 1-5, the standard deviation of ratings across all competitors is 1.5. To enable normalization and address the factors mentioned above, we also considered the ratings from Frequent Users those users who have had a minimum of two conversations with a particular socialbot. Using Frequent-Users ratings for CUX, we reduce variability across conversations and select into a well-calibrated set of interactors. Table \ref{conversation-statistics}  provides the information on  Alexa Prize conversation and rating distribution.

\vspace{-0.015\textheight}

\begin{table}[htb]
  \caption{Conversation Data and Ratings Statistics}
  \label{conversation-statistics}
  \centering
  \begin{tabular}{llll}
    \toprule
    Variable  & Counts and Ratings \\
    \midrule
    Total number of conversations  & Millions*     \\
    Total number of turns & Tens of Millions*       \\
    Average number of turns per conversation & 12      \\
    \noalign{\vskip 2mm} 
    \hline
    \noalign{\vskip 2mm}
    Counts of ratings from all Alexa Users   & Hundreds of Thousands* \\
    Counts of ratings from Frequent Users & Hundreds of Thousands* \\
    Average of all Alexa user ratings  & 3   \\
    Average of Frequent Users ratings  & 2.8  \\
    Average Engagement Evaluator ratings  & 2.4  \\
    \bottomrule
  \end{tabular}
  \vspace{-0.001\textheight}
    \begin{tablenotes}
      \small
 \item{*Rough Ranges}
    \end{tablenotes}
\end{table}


\vspace{-0.545cm}
\subsection{Engagement} \vspace{-0.01\textheight}
\vspace{-0.003\textheight}

To enable an open-ended, multi-turn conversation, engagement is
critical. Engagement is a measure of interestingness in a
conversation~\citep{yu2004detecting}. To account for this, we identify
proxies for engagement in our matrix of metrics for conversation
evaluation. We consider number of dialogue-turns and total conversation
duration an indicator of how engaged a user is in the conversation. We
recognize that there may be cases that may have a higher number of
turns due to inability of a socialbot being able to understand the user’s
intent, leading to follow-up turns with clarifications and
modifications, also potentially resulting in frustration at
times. However, analysis of a random sampling of conversations leads us to believe that the impact of this effect is negligible. To handle such
cases, we recruited a set of Alexa users (Engagement Evaluators) to rate their
conversations based on engagement. We were able to recruit over 2,000
Engagement Evaluators who scored over 10,000 conversations. Table \ref{conversation-statistics}
provides the mean of Engagement Evaluators Ratings (EER) with the
socialbots over the course of a month. The mean EER is significantly
lower than the mean all Alexa user ratings. We hypothesize
that when users tend to scrutinize the socialbots explicitly on engagement
and interestingness, they may rate conversations lower than if they
were rating overall experience, following the standard rating protocol.

\vspace{-0.3cm}
\subsection{Coherence} \vspace{-0.01\textheight}
\vspace{-0.003\textheight}
A coherent response indicates a comprehensible and relevant response
to a user’s request. A response can be deemed weakly coherent if it is somewhat related. For example, when a user says: "What do
you think about the Mars Mission?"; the response should be about the
Mars Mission, space exploration more broadly or something related. A
response related to Space Exploration but not exactly an opinion or something
related to politics, would be considered weakly coherent. For open-domain conversations, the complexity in the response space makes
this problem extremely hard. To capture coherence, we annotated
hundreds of thousands of randomly selected interactions for incorrect, irrelevant or inappropriate responses. Using the annotations, we calculated the
response error rate (RER) for each socialbot, as defined by:

\vspace{-0.02\textheight}

\begin{align*}
RER &= \frac{Number \thinspace of \thinspace incoherent \thinspace responses}{Total \thinspace number \thinspace of \thinspace utterances} 
\end{align*}

\vspace{-0.3cm} 
\subsection{Domain Coverage} \vspace{-0.01\textheight}
\vspace{-0.003\textheight}

A domain specific conversation agent may be more akin to goal-directed conversations, where the output response space is bounded. An agent which is able to
interact on multiple domains can be
considered more consistent with humans expectations. 
To account for this, we evaluated domain coverage for Alexa Prize by identifying the distribution of domains on Alexa Prize conversations for each socialbot. We identified the topic for each user utterance and responses across all the conversations for each socialbot. For example, "who is your favorite musician?" is classified into the music domain.. 

We calculated the entropy measure (degree of randomness) across distribution of number of conversations across different domains. 
A conversation was classified into a particular domain based on the domain capturing the maximum number of consecutive turns on a domain in the conversation. Entropy across domains enables us to understand if a socialbot has a
normal or biased distribution across those domains. A high
degree of entropy indicates breadth of coverage across many domains, as opposed to a lower value which indicates a narrower focus
on certain topics or domains. We also measured the standard deviation
(STD) of ratings across the five Alexa Prize domains
(Sports, Politics, Entertainment, Technology, Fashion).  To identify whether a socialbot has a biased rating distribution for
certain domains or performs equally well across all of them.  For evaluation, we optimized for high entropy while minimizing the standard deviation of the entropy across multiple domains. High entropy ensures
that the socialbot is talking about a variety of topics while a low standard
deviation gives us confidence that the metric is a confident one. To
combine entropy and standard deviation in ratings across domains for a socialbot,, we looked at Reverse
Coefficient of Variation (R-COV)~\citep{cov} as a comprehensive metric to evaluate domain
coverage. R-COV is obtained by taking the ratio of mean domain distribution based entropy across the conversations for each socialbot and corresponding standard deviation. 

\vspace{-0.3cm} 
\subsection{Conversational Depth} \vspace{-0.01\textheight}
\vspace{-0.003\textheight}

Coherence is usually measured at turn level. However, in a multi-turn
conversation, context may be carried over multiple turns. While
evaluating conversational agents, it is important to detect the
context and the depth of the conversations. Human conversations generally go deeper about a particular topic. An agent that is able to capture topical depth may sound more natural. To evaluate the
agents on conversational depth, we used the topical model to identify
the domain for each individual utterance. Conversational depth for a socialbot was calculated as the average of the number of consecutive turns on the same topical domain. 

\vspace{-0.3cm}
\subsection{Topical Diversity/Conversational Breadth} \vspace{-0.01\textheight}
\vspace{-0.003\textheight}

A good conversational agent is capable of: (i) identifying the topics and keywords from a given utterance (ii) able to have conversations around the same topics and (iii) can share related concepts (iv) identification of appropriate intent. Natural conversations are highly topical and humans frequently use keywords in their interactions. Agents lacking topical diversity might frustrate some users who are
not interested in the limited set of topics offered by the
socialbot. Evaluating conversational breadth is important to understand how
broadly an agent is able to converse as opposed to potentially having
user-pleasing but potentially highly-scripted conversations about a
small number of topics. As mentioned above, breadth depends on coarse topical domains
(e.g. Politics, Sports, Music, etc.) as well as fine-grained topical
keywords (e.g. Obama, Federer, John Lennon, etc.). We use topical
vocabulary size as a proxy for a signal on topical diversity. We also
measure the distribution of each topic for a socialbot which we use to
measure topic affinity for a socialbot.


\vspace{-0.3cm}
\subsection{Unification of Evaluation Metrics} \vspace{-0.01\textheight}
\vspace{-0.003\textheight}
Users tend to mentally evaluate the conversational systems at a more
fine-grained level. In five separate user studies,
we asked users to rate conversations with three socialbots on a
scale of 1 to 5. We learnt that even though users evaluated multiple
socialbots with same score, they had a clear rank order among the
socialbots, indicating that we need fine-grained information to systematically compare and evaluate the conversational agents. Conversational agents can be
evaluated on multiple dimensions, and agents may perform well on some,
and poorly on others. The proposed matrix of metrics should be unified
in a manner appropriate to the requirements. For Alexa Prize, we planned to generate a conversation-quality based ranking for the socialbots. 

We explored three strategies: stack ranking (with and without weights), winners circle, and confidence bands. For stack ranking, we rank the bots
on individual metrics and generate a score using a summation across
the metrics. A weighted stack ranking approach can be adopted if all the metrics are not equally important. However, if error bars in metric values indicate that the differences are not significant, stack ranking may not provide the most appropriate solution. To account for error bars, we tried the winners circle and confidence band approaches. For each metric, we defined the "winners circle" as all the socialbots that came within error bar (95\% confidence) of the "winners" (the overall top two performing bots as determined by Alexa user ratings). A socialbot within error bar of these two top socialbots was given a score of 1 (including the top socialbots) for that criteria, else it received a 0. An aggregate score was generated across all the evaluation metrics and 4 bands of socialbots emerged. Table \ref{unification-example}  provides an example in which bot 1 and bot 2 are the winners based on user ratings. Finally, we tried the "confidence bands" approach where a score of 1 to any bot within the 95\% confidence band of the top two socialbots for each individual metric (instead of the top user-rated socialbots being determined as the benchmark across all metrics).

\begin{table}[htb]
\begin{threeparttable}
  \caption{Sample Unification of Evaluation Metrics for Alexa Prize}
  \vspace{-0.005\textheight} 
  \label{unification-example}
  \centering
  \begin{tabular}{llllllllll}
    \toprule
    Metric & bot 1 & bot 2 & bot 3 & bot 4 & bot 5 & bot 6 & bot 7 & ... \\
    \midrule
    CUX: Mean User Rating & 1	& 1& 	1 &	0	& 1 &	0 &	0 &	...\\
    CUX: Mean Frequent-User Rating & 1 &	1 &	1	& 0 &	1 &	1	& 0& ...\\
    Coherence: RER & 1&	1&	0&	0&	1&	1&	0	& ...		\\
    Engagement: EER & 1&	1&	0&	0&	0&	0&	0&		...\\
    Engagement: Median Duration & 1&	1&	0&	0&	0&		0&	0&	...\\
    Engagement: Median Turns & 1&	1&	1&	1&	1&	0&		1&	...\\
    Domain Coverage: R-COV & 1&	1&	0&	1&	0&	1&		0&	...\\
    Topical Diversity: Vocab Size & 1&	 1&	1&	1&	0&	0&		0&	...\\
    Topical Diversity: Mean Freq & 1&	 1&	0&	0&	0&	0&		1&	...\\
    Conv. Depth: Mean Depth & 1&	1&	1&	1&	0&	0&		0&	...\\

   \hline 
    \textbf{Total Score} & 10 & 10 & 5 & 4& 4 & 3 & 2  & ...\\
       \bottomrule
   \end{tabular}
      \begin{tablenotes}
      \small
 \item{CUX: Conversational User Experience, RER: Response Error Rate, EER: Engagement Evaluator Rating, R-COV: Reverse Coefficient of Variantion}
    \end{tablenotes}
 \end{threeparttable}
 \vspace{-0.012\textheight} 
\end{table}

\vspace{-0.3cm}
\subsection{Automating User Ratings} \vspace{-0.01\textheight}
\vspace{-0.003\textheight}

In the current study, we use the ratings obtained from Alexa users as our ground
truth. To address this concern, we did a preliminary analysis on a
subset of data (about 60,000 conversations) for automation of user
ratings using utterance level and conversation level features. The following utterance level and conversational level features were used: N-grams of user-bot turns,
token overlap between user utterance and socialbot response, duration of the conversation, number of turns and mean response
time. We trained a model using Gradient Boosted Tree (GBDT) ~\citep{elith2008working} and Hierarchical LSTM (HLSTM)~\citep{serban2016building} to estimate user ratings of conversations. 

Moreover, in current study, the Coherence, Engagement and Conversational User Experience metrics are obtained by keeping humans in the loop. However, with the amount of data collected, it is possible to automate this process with supervised training. Techniques adopted by ~\cite{lowe2017towards} and ~\cite{adversarialTesting} and their variations can be further expanded to obtain these specific metrics apart from the general automation of the ratings mentioned above. 

\vspace{-0.4cm}
\section{Results and Discussion} \vspace{-0.01\textheight}

\subsection{Relevance of Evaluation Metrics}\vspace{-0.01\textheight}
\vspace{-0.003\textheight}

While aggregating evaluation metrics to come up with a unified metric, it is important to find the relevance of each of these metrics. For the Alexa Prize Competition, we evaluated the relevance by identifying the correlation of each of these metrics with user ratings and Frequent-User ratings. Table \ref{corelation-evaluation} provides the correlation with all Alexa user ratings, Frequent-User ratings and Engagement Evaluator ratings with evaluation metrics. In this section we share our findings. 

\begin{table}[htb]
\vspace{-0.01\textheight}
  \caption{Correlation of Evaluation Metrics with User Ratings}
  \label{corelation-evaluation}
  \centering
  \begin{tabular}{lllll}
    \toprule
    Metric & Users & Frequent-Users & Engagement Evaluators  \\
    \midrule
    CUX & 0.93 & 1 & 0.91 \\
    Coherence: RER & -0.88 &  -0.88 & -0.82 \\
    Engagement: EER & 0.93 & 0.91 & 1\\
    Engagement: Median duration & 0.81 & 0.82 & 0.77 \\
    Conversational Depth  & 0.73 & 0.73 & 0.80 \\
    Topical Diversity: Vocab. Size.  & 0.07* &  0.05* & 0.10* \\
    Topical Diversity: Topic Freq. & 0.37* & 0.25* & 0.42 \\
    Domain Coverage: R-COV & 0.24* &  0.23* & 0.40\\
    \bottomrule
  \end{tabular}
  \begin{tablenotes}
      \small
 \item{*P-value greater than level of significance: 0.05}
    \end{tablenotes}
 \vspace{-0.03\textheight}
\end{table}
\vspace{-0.01\textheight}
 
We found through user studies and data analysis that we need fine-grained information
to systematically compare and evaluate the conversational agents despite availability of user ratings. Fine-grained analysis also provides an insight on the areas of strengths and weakness of socialbots. We also compared the average ratings
provided by Alexa users and those provided by Frequent Users and Engagement Evaluators. From Table \ref{conversation-statistics}, it is clear that the ratings
provided by Frequent-Users and engagement evaluators is 7\% and 20\% lower than those provided by general
users (3.0 vs 2.80 vs 2.40) respectively. 

While optimizing for user experience (as measured by the rating mechanism proxy) is important to the Alexa Prize, the goal of this research is to create a more comprehensive conversational evaluation metric that enables improvement in individual component areas key to the overall human perception of successful conversation. We obtained the correlations of other metrics with user ratings and combined these metrics to create an evaluation model.

\textbf {Conversational User Experience:}  To capture subjectivity involved with CUX, as mentioned earlier we used Frequent-User ratings as a measure of evaluation. CUX is highly correlated with user ratings and EER (Table \ref{corelation-evaluation}). As a part of future work, we plan to train a model to predict CUX based on Frequent-User ratings seperately. As described earlier, ratings from Frequent-Users are capable of measuring subjectivity, expectation and other CUX metrics. It can be observed from the fact though these users come from the same pool of all Alexa users, still their average ratings are lower than average of all Alexa user ratings.

\textbf {Engagement:} To evaluate user engagement, we aggregated the ratings provided by engagement evaluators. Although the correlation between ratings obtained from all Alexa users and engagement evaluators is high (0.90), the average ratings provided by these evaluators is 20\% lower than the all Alexa user ratings. High correlation between all Alexa user ratings and EER implies that the user ratings can partially capture engagement. Other components for engagement that we considered are median duration and number of turns. We see a high correlation for median duration and number of turns for both Alexa user ratings and Frequent User ratings, indicating that longer conversations have a higher probability of higher ratings. While duration is not a measure of the information in a conversation it does provide insights on user satisfaction and engagement.

\textbf {Coherence:} For coherence, we evaluated response error rate (RER). We see a high negative correlation with RER (Table \ref{corelation-evaluation}), leading to the conclusion that users give poor ratings if responses are incoherent. 

\textbf {Conversational Depth:} We observed a high correlation between conversational depth for both all user ratings and frequent users implying deeper conversations tend to result in higher ratings from users.

\textbf {Topical Diversity:} We observed a positive directionality in correlation between the average frequency of topics and ratings, although the p-value is not very low. However, the correlation is high when evaluated with Engagement Evaluators (0.42), with low p-value. It can be hypothesized that when users are explicitly asked to engage with the socialbots, they tend to look for diversity in topics. If socialbots are able to respond with diversity in topics, if leads to higher user ratings. Another aspect of Topical Diversity we considered is the size of the topical vocabulary. Although the correlation between ratings and vocabulary is directionally positive, the p-value is insufficient to reach a clear conclusion. 

\textbf {Domain Coverage:} We used Reverse Coefficient of Variation (maximizing entropy minimizing standard deviation) to obtain this metric. We observed a weakly positive correlation with user ratings and Frequent User ratings. However, we found a high correlation (0.40 with low p-value) between R-COV and user ratings. Similar to Topical Diversity, it can be hypothesized that when users are explicitly asked to engage with the socialbots, they tend to cover broader domains and the socialbots that are able to respond appropriately tend to receive higher ratings.

Based on the analysis and observations, it can be concluded that the proposed metrics correlates strongly with ground truth. Hence these metrics can be used as measure for evaluating conversational agents. Given that ratings are obtained by keeping humans in the loop, which is not generally possible at scale, models can be trained to enable automated evaluation of conversational agents.

\vspace{-0.3cm}
\subsection{Unification of Evaluation Metrics} \vspace{-0.01\textheight}
\vspace{-0.003\textheight}

As mentioned in Section 4.2-4.7, it is important to unify the evaluation metrics mentioned above to be able to compare conversational performance in totality. For the Alexa Prize competition, we obtained the scores for each socialbot based on the unified metric, as exemplified in Table \ref{unifiedExample}. We found the correlation between the scores obtained by unified metric with user ratings and Frequent-User ratings.

\begin{table}[htb]
  \caption{Unified Metric - User Ratings Correlation}
  \label{unifiedExample}
  \centering
  \begin{tabular}{llll}
    \toprule
     & User Ratings & Frequent-User Ratings\\
    \midrule
    Correlation & 0.66  & 0.70 \\
    \bottomrule
  \end{tabular}
  \vspace{-0.01\textheight}
\end{table}

\vspace{-0.3cm}
\subsection{Automating User Ratings}\vspace{-0.01\textheight}
\vspace{-0.003\textheight}

We did a preliminary analysis on 60,000 conversations and ratings and we trained a model to predict user ratings. We observed the Spearman and Pearson correlations of 0.352 and 0.351 respectively (Table~\ref{corelation-stats})  with significantly low p-value with a model trained using Gradient Boosted Decision Tree. Although the results for GBDT are significantly better than Random selection for 5 classes and the model trained using Hierarchical LSTM, there is a need to extend this study the millions of Alexa Prize conversations. Furthermore, some of the evaluation metrics (coherence, topical depth, topical breadth, domain coverage, etc.)  obtained at conversation level can also be used as features. With significantly higher number of conversations combined with topical features, we hypothesize that the model would perform much better than the results obtained in preliminary analysis in Table~\ref{corelation-stats}. Given subjectivity in ratings, we appropriately found inter-user agreement to be quite low for ratings analysis. A user might give a conversation 5 stars because he/she thought the socialbot was humorous, while another user might find it unknowledgeable. Users may have their own criteria to evaluate the socialbots. Therefore, as a part of the future work, we would like to train the model with user level features as well. This experiment was done to obtain the potential of automating the ratings.

\begin{table}[htb]
\vspace{-0.01\textheight}
  \caption{Correlation of the regression model with user ratings}
  \label{corelation-stats}
  \centering
  \begin{tabular}{llll}
    \toprule
    Algorithm & RMSE & Spearman & Pearson \\
    \midrule
    Random & 2.211& 0.052 & 0.017\\
    HLSTM & 1.392 & 0.232 & 0.235\\
    GBDT & 1.340& 0.352 & 0.351\\
    \bottomrule
  \end{tabular}
  \vspace{-0.02\textheight}
\end{table}










\vspace{-0.2cm}
\section{Conclusion and Future Work} \vspace{-0.01\textheight}
\vspace{-0.005\textheight}

Evaluating open-domain conversational agents is a challenging task and
has remained largely unsolved. In this work, we defined various metrics
which can be used to evaluate open-domain conversational agents and
proposed a mechanism to obtain those metrics. We have used these
metrics to evaluate the open-domain conversational agents (socialbots) built for
Alexa Prize, a university competition targeted towards advancing
the state of Conversational AI. The challenge is to build a socialbot which can converse coherently and engagingly on popular topics and current events for 20 minutes with humans. During the competition, we have obtained millions
conversations and corresponding ratings from Alexa users. After each
conversation, Alexa users are asked to give a rating and feedback,
which are currently considered as a baseline for us to evaluate our
metrics. We proposed the following metrics to evaluate the open-domain
agents: Conversational User Experience, Coherence, Engagement, Domain
Coverage, Topical Depth and Topical Diversity. We have also proposed a
mechanism to unify these metrics to obtain a single metric for
evaluation and comparison. Strong correlation between the unified
metric and user ratings indicate that we can use the unified metric as a
proxy to user ratings. To our knowledge, it is the largest evaluation to date of user ratings for conversational agents, featuring millions of conversations and hundreds of thousands of ratings from Alexa users. We also present a preliminary analysis on building models, using 60,000
conversations to automate rating prediction with promising results. As
a part of future work, we plan to extend the preliminary work by
incorporating a significantly larger dataset. This model will also be helpful in
improving and evaluating various dialogue strategies automatically and
reliably. We would also like to test scalability of some metrics from previous work done on smaller datasets to see if they can be incorporated into the process. 

\vspace{-0.01\textheight}

\subsubsection*{Acknowledgments}
\vspace{-0.01\textheight}
We would like to thank all the university students and their advisors
(\cite{alexaPrizeTeams} )who participated in the competition. We
would also like to thank the entire Alexa Prize team (Science, Engineering, User Experience, Marketing, Legal, Program Management, and Leadership) for their useful suggestions and assistance during
the process. We would also like to thank Tagyoung Chung for the helpul edits.



\small
\bibliographystyle{acl_natbib}
\bibliography{mybib}



  

\end{document}